\documentclass[lettersize,journal]{IEEEtran}
\usepackage{amsmath,amsfonts}
\usepackage{algorithmic}
\usepackage{algorithm}
\usepackage{array}
\usepackage[caption=false,font=normalsize,labelfont=sf,textfont=sf]{subfig}
\usepackage{textcomp}
\usepackage{stfloats}
\usepackage{url}
\usepackage{verbatim}
\usepackage{graphicx}
\usepackage{cite}
\usepackage{booktabs}
\usepackage{multirow}
\usepackage{color}
\usepackage{tcolorbox}

\begin{document}

% \title{Post-Semantic-Thinking: A Robust Strategy to Distill Reasoning Capacity from Large Language Models}
\title{Distilling Reasoning Ability from Large Language Models with Adaptive Thinking
%Adaptive Thinking Mechanism: Distill Reasoning Ability from Large Language Models Based on Perceiving Task Complexity
}
\author{Xiaoshu Chen, Sihang Zhou, Ke Liang, Xinwang Liu, ~\IEEEmembership{Senior~Member,~IEEE}
        % <-this % stops a space
\IEEEcompsocitemizethanks{
\IEEEcompsocthanksitem Xiaoshu Chen, Ke Liang, and Xinwang Liu  are with the College of Computer Science and Technology, National University of Defense Technology, Changsha 410073, China. Sihang Zhou is with the College of Intelligence Science and Technology, National University of Defense Technology, Changsha 410073, China. \emph{Corresponding Author: Xinwang Liu.}, e-mail: xinwangliu@nudt.edu.cn
\IEEEcompsocthanksitem Manuscript Accepted July 14, 2025;
}
}

% \markboth{Journal}%
\markboth{IEEE Transactions on Neural Networks and Learning Systems}%
{Shell \MakeLowercase{\textit{et al.}}: A Sample Article Using IEEEtran.cls for IEEE Journals}

% \IEEEpubid{0000--0000/00\$00.00~\copyright~2024 IEEE}

\maketitle

\begin{abstract}
Chain of thought distillation (CoT-distillation) aims to endow small language models (SLM) with reasoning ability to improve their performance towards specific tasks by allowing them to imitate the reasoning procedure of large language models beyond simply predicting the answers. Most existing CoT-distillation methods adopt a pre-thinking mechanism, allowing the SLM to generate a rationale before answering. In this way, pre-thinking enables SLM to analyze questions but makes answer correctness sensitive to minor errors in rationale. Therefore, we propose a robust post-thinking mechanism to generate answers before rationale. Thanks to this answer-first setting, 1) the answer can escape from the rationale-sensitive problem; 2) the rationale serves as an error amplifier, making SLM focus on learning hard samples; and 3) the inferring efficiency can also benefit. Although post-thinking brings many advantages, it may lose the ability to analyze complex questions compared to pre-thinking. Therefore, a plug-and-play adaptive-thinking mechanism is proposed to integrate the merits of pre-thinking and post-thinking, in which a perception module based on soft prompt tuning is introduced to prompt SLM to answer or think first according to the complexity of questions. Extensive experiments are conducted across 12 datasets and 2 language models to demonstrate the effectiveness of the proposed mechanism.
\end{abstract}


\section{Introduction}
\IEEEPARstart{C}hain of thought (CoT) \cite{chu2023survey}, as an effective strategy, has been widely used to enhance the performance of large language models (LLM) \cite{brown2020language, touvron2023llama, llama2} on various complex reasoning tasks. However, its usage is still limited since it is observed that such benefits on reasoning ability brought by CoT only exist in LLM with more than 100B parameters but not in smaller models \cite{wei2022chain}.

To alleviate such a problem, prior attempts \cite{ho-etal-2023-large, hsieh-etal-2023-distilling, li2023turning, renzhu2022specializing, zhu2023pad} adopt chain of thought distillation (CoT-distillation) to transfer reasoning capacity from teacher LLM to student SLM, thereby improving the student SLM's performance in the vertical field such as complex question answering \cite{liu2022disentangled}, recommender systems \cite{10506571}, knowledge graphs reasoning \cite{Liangke_Survey, 9416312} and LLM-based Agents \cite{xi2023rise}. 

\begin{figure*}[tb]
\centering
\setlength\abovecaptionskip{-0.05cm}
\setlength\belowcaptionskip{-1cm} 
\includegraphics [width = 14cm] {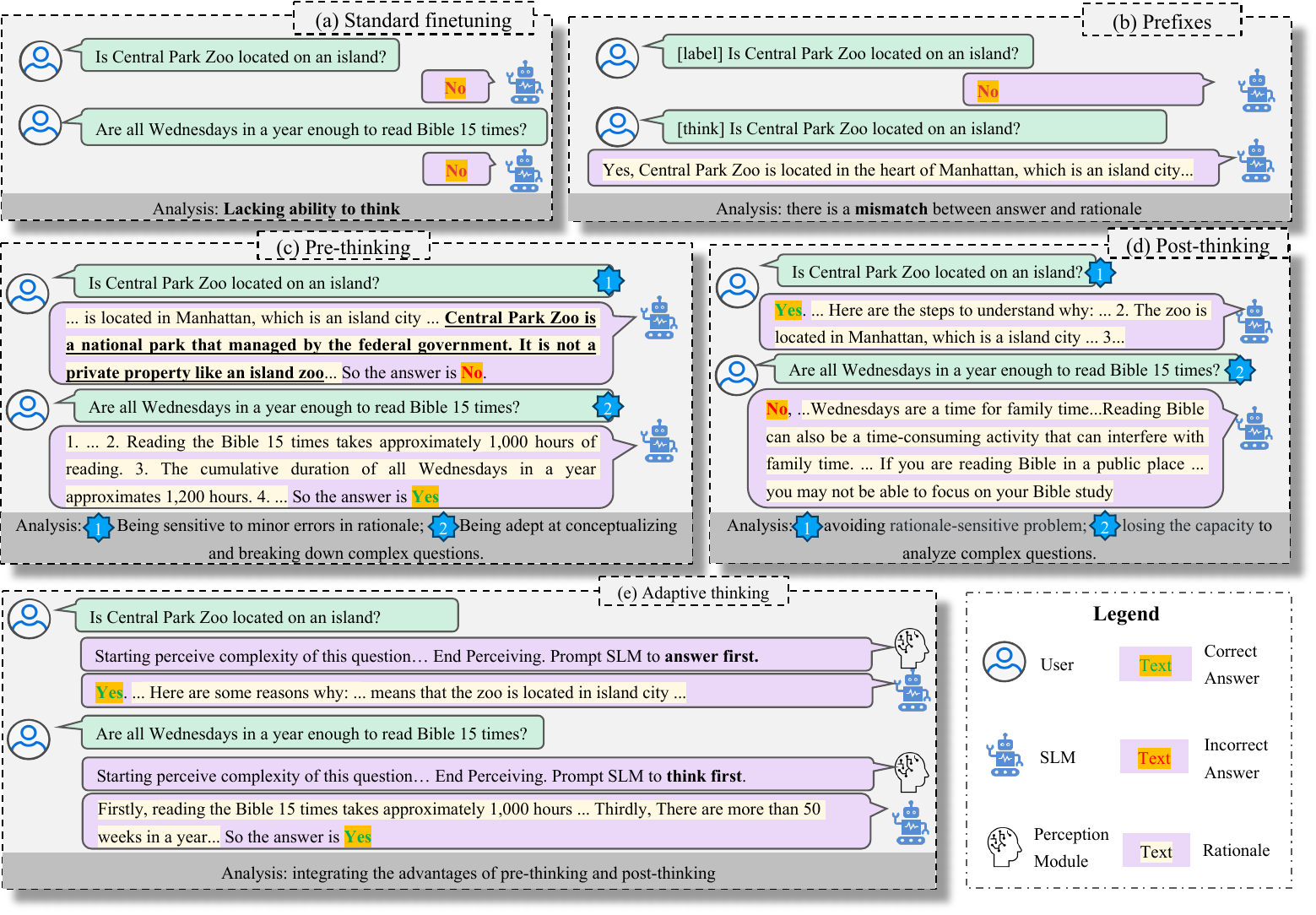}
\caption{Comparison of different CoT-distillation mechanisms. Standard finetuning trains an SLM to generate only the answer to the question. Prefixes train SLM to complete the task of generating answers or rationale under the instruction of different prefixes. Pre-thinking makes SLM think and analyze before answering while post-thinking lets SLM answer first. Through the proposed perception module, ATM allows SLM to adaptively decide whether to pre-think or post-think based on the complexity of the problem.}
\label{fig1}
\end{figure*}

There are two main CoT-distillation paradigms currently. One is the prefixes mechanism \cite{hsieh-etal-2023-distilling}, and the other is the pre-thinking mechanism \cite{ho-etal-2023-large,magister2022teaching}. As shown in Fig.\ref{fig1}.(b), since rationale and answer are not in the same generation sequence with prefixes mechanism, the contextual association between them is severed, resulting in sub-optimal performance and that rationale mismatches the answer. Thus, pre-thinking, which generates rationales before answers, becomes the most developed CoT-distillation mechanism.

% Compared with standard finetuning that only uses the golden answer to train SLM, both of these two cot-finetuning paradigms make SLM imitate the reasoning procedure (also referred to as rationale) of teacher LLM beyond simply predicting the answer during training to elicit the reasoning capacity of student SLM. The difference between them is that the former applies multi-task learning that distinguishes tasks by prefixes, while the latter uses rationale and answer as a sequence generation target. As shown in Fig.\ref{fig1}.(b), since rationale and answer are not in the same generation sequence with prefixes mechanism, the contextual association between them is severed, resulting in sub-optimal performance and that rationale mismatches the answer. Thus, pre-thinking is the most developed cot-finetuning mechanism currently. 

With the aid of the pre-thinking mechanism, the SLM acquires thinking and reasoning capabilities for complex questions, akin to those exhibited by LLMs. However, a critical issue, termed the rationale-sensitive problem, arises in pre-thinking and cannot be ignored. Specifically, under the next token prediction (NTP) training objective, the rationale-answer generation sequence causes answer generation to become excessively reliant on the rationale, rendering the correctness of answers highly sensitive to minor errors in the rationale. Consequently, as illustrated in the output of the SLM for the first question in Fig.\ref{fig1}.(c), although the initial part of the rationale contains correct information, the answer is rendered incorrect due to subsequent analysis (highlighted in bold) deviating from the intended direction. Notably, minor errors in rationale are a pervasive issue for SLM.

To solve this problem, we propose a robust post-thinking mechanism. In the post-thinking mechanism, the rationale-answer generation sequence is restructured into an answer-rationale sequence. Since the answer is generated prior to the rationale, the rationale primarily serves as an explanation for the answer, enabling the student SLM to avoid the adverse effects of minor errors in the rationale. Moreover, as the answer can be generated without the need for a rationale, post-thinking demonstrates higher efficiency compared to pre-thinking during inference. Last but not least, the post-thinking mechanism possesses a notable advantage: its inherent ability to focus on learning from hard samples. When the SLM produces an incorrect answer, it attempts to generate an unusual rationale to justify the error. As illustrated in the rationale provided by the SLM for the second question in Fig.\ref{fig1}.(d), the SLM attempts to justify the incorrect answer through irrelevant perspectives, such as conflicts with family time or distractions in a public place. This enforced interpretation in post-thinking, guided by incorrect answers, causes the generated rationale to deviate further from the correct interpretation compared to that produced by pre-thinking, thus leading to greater loss on incorrect samples. This greater loss during training will drive the SLM to concentrate more on learning from the corresponding samples.

However, while post-thinking enables the SLM to focus on mastering hard samples, it is impractical to achieve proficiency in addressing highly complex questions solely by intensifying learning attention without altering the problem-solving approach. The second question in Fig\ref{fig1}.(d) represents a typical hard sample, distinguished from ordinary hard samples by its higher complexity. For this question, the SLM cannot derive the answer using a single piece of knowledge, as it does for the first question in Fig.\ref{fig1}.(c). Instead, the SLM must decompose it into two sub-questions: (1) how many Wednesdays are there in a year, and (2) how long it takes to read the Bible 15 times. Clearly, this type of high-complexity hard sample is better suited for pre-thinking, which can decompose a question into sub-questions through reasoning. 

Human cognitive processes can be broadly categorized into fast and slow thinking: the former quickly generates answers based on intuition, while the latter engages in deeper reasoning and analytical processing \cite{fast_and_slow, fast_and_slow_1}. When answering questions, humans select between fast and slow thinking depending on factors such as prior experience, environmental context, and question complexity, aiming to optimize both efficiency and effectiveness. Similarly, post-thinking and pre-thinking mirror characteristics analogous to fast and slow thinking, respectively. Building on this cognitive paradigm, we propose the adaptive thinking mechanism (ATM), which dynamically selects between pre-thinking and post-thinking based on the SLMs' capabilities and the complexity of a given question.

Specifically, we incorporate a perception module into the ATM, which evaluates the complexity of input questions and generates a sequence of soft prompt tokens accordingly. These soft prompt tokens are then prepended to the input tokens to guide the SLM in determining whether the questions require reasoning or can be answered directly. Owing to this perception module, the ATM not only retains the pre-thinking capability to reason through and decompose complex questions but also, aided by post-thinking, mitigates the rationale-sensitive problem and enhances the focus on learning hard samples. Furthermore, as the ATM is orthogonal to pre-thinking mechanisms, it can also serve as a plug-and-play enhancement for most methods based on pre-thinking. 

The main contributions of this paper are as follows:

\textit{1) Post-thinking mechanism}: a simple yet effective and efficient mechanism that adjusts the rationale-answer generation sequence to the answer-rationale generation sequence, thus escaping from the rationale sensitivity problem and improving attention on learning hard samples.

\textit{2) Adaptive thinking mechanism}: a plug-and-play framework that adaptively selects between pre-thinking and post-thinking based on the complexity of the question, effectively integrating the strengths of both approaches.

\textit{3) Better performance}: extensive ablation experiments across 12 tasks with two publicly available models (GPT2 and T5) demonstrate the effectiveness of ATM relative to the prefixes mechanism and pre-thinking mechanism.

\section{Related Works}
\subsection{Chain of Thought Distillation}
CoT is first proposed by Wei et al \cite{wei2022chain} to enhance the model’s reasoning capacity in various complex reasoning tasks. Subsequent attempts optimize CoT in the following three directions: constructing more effective CoT prompt \cite{gao2023pal, chen2022program, imani-etal-2023-mathprompter, lu2023dynamic, zhang2023automatic}, modeling more complex reasoning with tree \cite{long2023large,yao2023tree} or graph \cite{lei2023boosting, besta2023graph} structure, enhancing reasoing with external knowledge, tools, and post-processing strategy \cite{lu2023chain, wang2023knowledge, wang2023selfconsistency,DBLP:journals/corr/abs-2110-14168}.

However, CoT can only improve the reasoning capacity of LLM but not that of SLM \cite{wei2022chain}. Therefore, a lot of works introduce knowledge distillation \cite{HintonVD15} to transfer the reasoning capacity of LLM to SLM by having SLM study the more informative output of LLM. Different from the previous widely used soft distillation methods that align the distribution of the small model to the large model in the output \cite{LiuDistilling}, attention \cite{WangMINILM}, and hidden states \cite{sunetal2019patient}, the distillation for the reasoning capacity of LLM adopts the hard distillation that aligns the SLM's generation sequence with that of LLM.

There are two main mechanisms for this hard distillation. 
\subsubsection{Prefix mechanism} Distilling step by step \cite{hsieh-etal-2023-distilling} is a pioneer of the prefix mechanism, which applies task prefixes to distinguish whether the model generates an answer or rationale. Based on it, Mind's Minor \cite{minds} introduces another self-evaluation task to further improve model performance and Mint \cite{liang-etal-2024-mint-boosting} explores the effect of prefix mechanism on mathematical reasoning. Although these methods retain the information of the rationale generated from LLM, it separates the contextual connection between the answer and the rationale, resulting in a mismatch between them. To alleviate the mismatch, MMI \cite{MMI} maximizes the mutual information between the answer and the rationale. However, since the problem of not being in the same context has not been fundamentally solved, the mismatch problem is still quite significant. 
\subsubsection{Pre-thinking mechanism} Basic pre-thinking treats the rationale-answer sequence as the target generation sequence for SLM \cite{ho-etal-2023-large,magister2022teaching}. Numerous methods have been proposed to enhance the performance of the basic pre-thinking mechanism. Multi-step reasoning \cite{FuSpecializing}, MD \cite{Potmixed}, and UniCoTT \cite{structureUni} combine various forms of data (e.g., different CoT structures) for training. DISCO \cite{chen-etal-2023-disco}, SCOTT \cite{wang-etal-2023-scott}, and MVC-CD \cite{counterdis} incorporate counterfactual distillation strategies into pre-thinking. Pad \cite{zhu2023pad} and MD \cite{Potmixed} integrate programming-aid capabilities. SGFT \cite{multistepSG} explores the effect of distilling solution guidance instead of complete rationales. KPOD \cite{structureUnikeypoint} proposes a progressive distillation strategy to help SLMs better comprehend the rationale provided by LLMs. The incorporation of negative data \cite{li2023turning} and multiple model integration \cite{shridhar-etal-2023-distilling} has also been shown to enhance the pre-thinking mechanism. 

However, while the aforementioned methods improve the ability of SLMs to learn the reasoning capacity of LLMs, the rationale-sensitive problems remain unresolved.
Thus, we investigate the third mechanism, called post-thinking to distill reasoning capacity from LLM to SLM. Furthermore, ATM is introduced to fully integrate the advantages of pre-thinking and post-thinking.
It is worth noting that there are several human-inspired dual-system thinking frameworks \cite{yu2025long, kang2025c3ot, liang2025thinkswitcher, yao2024hdflow, su2024dualformer}, comprising fast and slow thinking, to enhance the reasoning capabilities of language models. Compared to ATM, these methods exhibit two limitations: 1) their fast thinking component still relies on pre-generated rationales, albeit shorter in length; and 2) the SLM mimics external rules or LLMs' predictions, rather than being informed by its internal capabilities to chosse thinking mode, resulting in suboptimal performance.

\subsection{Soft Prompt Tuning}
Prompt \cite{gpt2} refers to words/sentences used to stimulate the capability of LLM in completing a certain task. As long as a good prompt can be found, LLM without fine-tuning can also achieve the performance as full-finetuing. However, manually optimizing these discrete natural language prompts to maximize LLM performance can be challenging. Therefore, researchers have introduced the concept of soft prompt tuning \cite{liu2023gptunderstand, li-liang-2021-prefix, lester-etal-2021-power}. In soft prompt tuning, the prompt is randomly initialized to a continuous embedding and then added to the original input. Typically, all parameters of LLM except the prompt are frozen, i.e., the goal of soft prompt tuning is to find the most appropriate prompt for LLM in an end-to-end manner.

Most soft prompt tuning methods optimize only the prompt embedding at the input layer of the model \cite{liu2023gptunderstand, li-liang-2021-prefix}. Unlike these methods, prefix-tuning \cite{lester-etal-2021-power} inserts prompt embedding into each transformer layer of the language model. In this way, prefix-tuning achieves closed results comparable to finetuning on various generation tasks using models with fewer than 1 billion parameters. In addition to these methods, recent advancements \cite{vu-etal-2022-spot, su-etal-2022-transferability, hambardzumyan2021warp} explore how to pretrain soft prompts or utilize prompts from different tasks to reduce the computational effort required for finetuning soft prompts for new tasks.

For our work, prefix-tuning is applied in the perception module to optimize the soft prompt that is used to prompt the SLM to pre-think or post-think. It is worth noting that, unlike the standard soft prompt tuning process where only the parameters related to the soft prompt are optimized, in ATM, the parameters of the language model are continuously optimized as well, because solely optimizing the soft prompt parameters often does not achieve the same effectiveness as finetuning all the parameters for SLM \cite{su-etal-2022-transferability}.

\begin{figure*}[tb]
\centering
\includegraphics [width = 12cm] {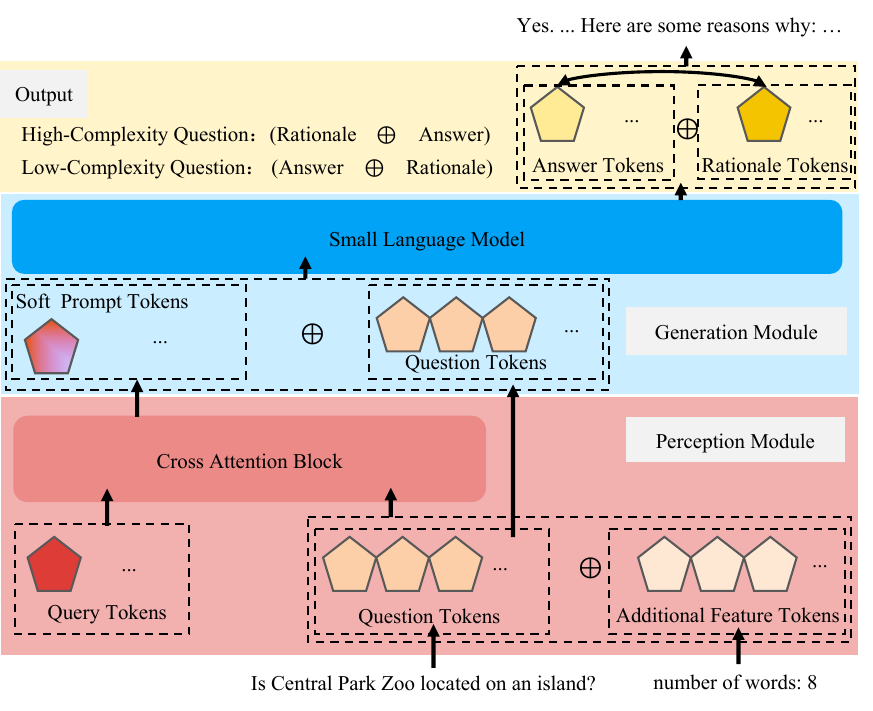}
\caption{Illustration of the ATM. The perception module perceives the complexity of the task and generates soft prompt tokens to prompt the generation module whether to answer or think first. The double arrow on the answer and rationale tokens indicates that the order of generating tokens can be swapped based on the complexity of the question. And the query tokens are the randomly initialized embeddings.}
\label{fig2}
\end{figure*}

\section{Methodology}
In this section, the post-thinking mechanism will first be elaborated in detail. Then, a systematic introduction on how to unify post-thinking and pre-thinking adaptively in ATM based on the soft prompt tuning will be presented. Fig.\ref{fig2} illustrates the overall architecture of the proposed ATM.

To better understand, we first formulate some concepts. Denote \(f_s\) to be the small language model (generation module in Fig. \ref{fig2}) with parameter \(\theta_s\) and \(f_p\) to be the perception module with parameter \(\theta_p\). \(S = \{(x_i, y_i, r_i)^{N}\}\) refers to a dataset with \(N\) training samples, where \(x_i\) and \(y_i\) represent a question and answer, respectively. And \(r_i\) is a rationale corresponding to \(x_i\), which is generated from LLM based on the zero-shot CoT prompt ``\textit{Let’s think step by step}''. Specifically, similar to previous work \cite{wang-etal-2023-scott}, we revise the CoT prompt to \textit{``Let's think step by step why the answer is \(y_i\)"}. In this way, LLM can focus on generating \(r_i\) with the prior information of \(y_i\), thereby reducing minor error in \(r_i\) and improving the consistency between \(y_i\) and \(r_i\).

\subsection{Post-Thinking Mechanism}
Since the post-thinking mechanism swaps the order of generating rationale and answer in the pre-thinking mechanism, the pre-thinking mechanism will be introduced first to compare them better. 

The training object of pre-thinking is to maximize the likelihood of the SLM to generate \(r_i \) and \(y_i\), which can be formulated as follows:
\begin{gather}
\mathbb{E}_{(x_i, r_i, y_i)\sim S}[logP(r_i | x_i; \theta_s) \cdot logP(y_i | x_i, r_i; \theta_s)]
\label{eq1}
\end{gather}
By examining Equation \ref{eq1}, it is evident that the generation of the answer relies on \(x_i\) and \(r_i\). Given that \(r_i\), produced by the LLM, typically contains the answer information, the answer generation by the trained SLM will be highly dependent on \(r_i\) rather than \(x_i\). Consequently, minor errors in \(r_i\) (even not errors, but just some irrelevant content), can significantly impact the accuracy of the answer. More critically, compared to LLM, SLM is more prone to generating erroneous information when generating rationales.

To overcome this obstacle, post-thinking adjusts the generation order of \(r_i \) and \(y_i\). The training object post-thinking can be formulated as follows:
\begin{gather}
\mathbb{E}_{(x_i, r_i, y_i)\sim S}[logP(y_i | x_i; \theta_s) \cdot logP(r_i | x_i, y_i; \theta_s)]
\label{eq2}
\end{gather}
Obviously, the answer generation by SLM trained through Equation \ref{eq2} will no longer rely on rationale. Moreover, compared to the rationale in pre-thinking, the rationale in post-thinking will enhance model performance in a different manner, functioning as an error amplifier. When SLM makes incorrect predictions, rationales, serving as explanations for the answers, will tend to argue for the correctness of erroneous answers from peculiar perspectives, thus amplifying the distributional difference between rationales generated by SLM and those by LLM. In other words, for samples where SLM predicts incorrectly, their loss function will be larger, leading SLM to focus more on learning from these challenging samples during training.

\begin{algorithm}
    \caption{Training process of the ATM}
    \begin{algorithmic}[1] %每行显示行号
        \REQUIRE Training data \(S=\{x_i, y_i\}^N\), training epochs \(n_e\)
        \STATE Get \(r_i\) from LLM using zero-shot prompt ``\textit{Let’s think step by step why the answer is} \(y_i\)"
        \STATE Train pre-thinking model with parameters \(\theta_{pre}\) using Equation \ref{eq1}
        \STATE Train post-thinking model with parameters \(\theta_{post}\) using Equation \ref{eq2}.
        \STATE Split \(S\) into \(S_{pre}\) and \(S_{post}\) using Equation \ref {eq5}
        \STATE Initialize parameters \(\theta_s\) and \(\theta_p\)
        \FOR{epoch in \(n_e\)}
            \WHILE{iteration is effective}
            
                \STATE Updata \(\theta_s\) and \(\theta_p\) with the Equation \ref {eq6}, \ref{eq7}
                
            \ENDWHILE
        \ENDFOR
        \ENSURE  \(\theta_s\) and \(\theta_p\).
    \end{algorithmic}
\label{algorithm}
\end{algorithm}

\subsection{Adaptive Thinking Mechanism}
Although post-thinking can solve rationale-sensitive problems, it may lose the explicit thinking process to the question before answering it, which is extremely important when SLM solves high-complexity questions.  Therefore, ATM is proposed inspired by the way humans answer questions. To be specific, we prepend a perception module to SLM (generation module) to perceive the complexity of the input question and output a soft prompt to prompt SLM whether to pre-think or post-think. Below, we first elaborate on the principles of the perception module and the generation module. Then, the training process of the entire ATM is detailed.

\subsubsection{Perception module} The perception module consists of a transformer attention layer whose hyperparameter configurations are consistent with the transformer layer in SLM. 

Firsty, it takes \(K\) randomly initialized but learnable token embeddings, termed query tokens \(q \in \mathbb{R}^{K\times H}\), and \((d = x^t_i \oplus a^t_i) \in \mathbb{R}^{L\times H}\) as input, where \(H\) and \(L\) are the hidden size of SLM and the length of \(d\) respectively, \(x^t_i\) and \(a^t_i\) represent the token embedding corresponding to \(x_i\) and additional feature tokens \(a_i\) respectively, and \(\oplus\) represents the concatenation of the sequence.  \(x_i\) inherently contains implicit complexity-related features, such as word count, linguistic complexity, readability, and the number of reasoning steps required. \(a_i\) is intended to reduce the learning difficulty of the perception module during training by providing additional explicit information. For convenience, only the word number of \(x_i\) whose specific form is ``\textit{number of words:  }\textit{num}'', is utilized as the feature in our main experiments, where \textit{num} is the specific word number of the input question. In practical tasks, adaptations to the features in \(a_i\) could be made to reduce the difficulty of perceiving input questions' complexity. 

Then, the cross-attention module is implemented to output the soft prompt tokens \(p \in \mathbb{R}^{K\times H}\). The specific process can be expressed by the following formula:
\begin{gather}
Q = qW_Q, K = dW_K, V = dW_V \\
p = Softmax(QK^T)V + q
\label{eq3}
\end{gather}
where \(W_Q \in \mathbb{R}^{H\times H}\), \(W_K \in \mathbb{R}^{H\times H}\) and  \(W_V \in \mathbb{R}^{H\times H}\) refer to the trainable parameter matrices. After the cross-attention module, the useful information in \(d\) for perceiving complexity of the question is gathered into \(p\) to help the perception module prompt the SLM.

\subsubsection{Generation module} Once \(p\) is generated, \(p \oplus x_i\) is fed into the language model to obtain the answer and rationale. Since the soft prompt tuning method used in this work is prefix tuning \cite{li-liang-2021-prefix}, \(p\) is also prepended to the input of each transformer layer of SLM. The language model can be any of the current mainstream generative language model frameworks, such as GPT \cite{gpt2} (decoder-only) or T5 \cite{t5} (encoder-decoder). Whether to generate an answer first or to generate a rationale first depends on the prompt contained in \(p\).

\subsubsection{Training procedure} First, we design a labeling assignment strategy to determine whether the input question should be thought first or answered first. According to Wan et al. \cite{wan2024knowledge} and Marion et al \cite{anonymous2023when}.\, the probabilistic distribution of the language model's output sequence can reflect its certain inherent knowledge in understanding the text. Based on this, we first train two SLMs with parameters \(\theta_{pre}\) and \(\theta_{post}\) using Equation \ref{eq1} and Equation \ref{eq2} respectively. It should be noted that a 5-fold cross-validation trick is adopted when training these two SLMs to prevent the overfitting of the model on the training set from affecting the label assignment. Then, \(x_i\), \(y_i\) and \(r_i\) is input into these two SLMs to obtain two output distributions, \(D^{\theta_{pre}}_{r_i \oplus y_i} \in \mathbb{R}^{B \times V}\) and \(D^{\theta_{post}}_{y_i \oplus r_i} \in \mathbb{R}^{B \times V}\), over \(r_i \oplus y_i\) and \(y_i \oplus r_i\), where \(B\) is the length of \(r_i \oplus y_i\) and \(V\) is the size of SLM's vocabulary. Subsequent, we construct two one-hot label matrices \(O_{r_i \oplus y_i} \in \{0, 1\}^{B \times V}\) and \(O_{y_i \oplus r_i} \in \{0, 1\}^{B \times V}\), where each row represents a one-hot vector of the corresponding golden token in \(r_i \oplus y_i\) and \(y_i \oplus r_i\). The distribution difference between output and one-hot label matrices can be calculated according to the following formula:
\begin{gather}
diff_{D,O} = \mathbb{D}(D, O)
\label{eq4}
\end{gather}
where \(\mathbb{D}(\cdot, \cdot)\) refers to the discrepancy function between the two input matrices. In this work, \(\mathbb{D}(\cdot, \cdot)\) is defined as cross-entropy to follow Marion et al \cite{anonymous2023when}. Finally, the following formula is applied to indicate whether the input question should be pre-thought or post-thought:
\begin{equation}
c_i =\begin{cases}
\text{1} & diff_{D^{\theta_{pre}}_{r_i \oplus y_i}, O_{r_i \oplus y_i}} < diff_{D^{\theta_{post}}_{y_i \oplus r_i}, O_{y_i \oplus r_i}}
\\
\text{0}& \text{otherwise}
\end{cases}
\label{eq5}
\end{equation}
where \(c_i = 1\) represents the question should be pre-thought and \(c_i = 0\) refers to the question should be post-thought. That is to say, when \(diff_{D^{\theta_{pre}}_{r_i \oplus y_i}, O_{r_i \oplus y_i}} < diff_{D^{\theta_{post}}_{y_i \oplus r_i}, O_{y_i \oplus r_i}}\), pre-thinking model can better understand the knowledge and logic in \(r_i\) for answering the question than post-thinking model, which reflects pre-thinking model is more suitable for solving the current question.

\begin{table*}[tb]
\begin{center}
\caption{The performance on Decoder-only Language model GPT2-Large. Accuracy (\%) is reported across 12 reasoning tasks with different methods.}
\renewcommand\arraystretch{1.5}
\label{tab1}
\renewcommand\arraystretch{1.5}
\resizebox{\linewidth}{14mm}{
\begin{tabular}{ c c c  c  c c  c  c c  c  c c  c  c c c}
\toprule[1.05pt]
Methods   & Pre   & Post   &\multicolumn{1}{c}{\begin{tabular}[c]{@{}l@{}}SE\end{tabular}} & \multicolumn{1}{c}{\begin{tabular}[c]{@{}l@{}}AS\end{tabular}} & \multicolumn{1}{l}{\begin{tabular}[c]{@{}l@{}}MA\end{tabular}} & \multicolumn{1}{c}{\begin{tabular}[c]{@{}l@{}}GSM\end{tabular}} & \multicolumn{1}{c}{Aqua}     & \multicolumn{1}{c}{Svamp}   & \multicolumn{1}{c}{\begin{tabular}[c]{@{}l@{}}DU\end{tabular}} & \multicolumn{1}{c}{\begin{tabular}[c]{@{}c@{}}TSO\end{tabular}} & \multicolumn{1}{c}{\begin{tabular}[c]{@{}l@{}}LLC\end{tabular}} & \multicolumn{1}{c}{\begin{tabular}[c]{@{}c@{}}CF\end{tabular}} & \multicolumn{1}{c}{\begin{tabular}[c]{@{}c@{}}CSQA\end{tabular}} & \multicolumn{1}{c}{\begin{tabular}[c]{@{}c@{}}SQA\end{tabular}} & \multicolumn{1}{c}{\begin{tabular}[c]{@{}l@{}}SQA-GLM4\end{tabular}}\\ 

\toprule[1.05pt]
Standard Finetuning&            &            &12.50&16.80&22.22&6.14&29.92&8.33&62.16&38.22&2.00&98.67&52.66&63.31 &63.31  \\
Pre-thinking       &$\checkmark$&            &12.50&14.28&20.55&6.97&29.52&9.33&36.03&39.11&2.66&\textbf{100.00}&34.97&63.61 &62.59\\
Post-thinking      &            &$\checkmark$&\textbf{13.81}&15.96&23.88&6.59&31.10&8.33&67.56&42.66&4.00&\textbf{100.00}&56.42&63.90  &63.60\\
ATM                &$\checkmark$&$\checkmark$&\textbf{13.81}&\textbf{18.48}&\textbf{25.55}&\textbf{7.66}&\textbf{31.89}&\textbf{10.33}&\textbf{72.07}&\textbf{43.55}&\textbf{4.66}&\textbf{100.00}&\textbf{56.67}&\textbf{65.06} &\textbf{64.48} \\
\toprule[1.05pt]
\end{tabular}}
\end{center}
\end{table*}

\begin{table*}[h]
\begin{center}
\caption{The performance on Encoder-Decoder Language model T5-Large. Accuracy (\%) is reported across 12 reasoning tasks with different methods.}
\renewcommand\arraystretch{1.5}
\label{tab2}
\renewcommand\arraystretch{1.5}
\resizebox{\linewidth}{14mm}{
\begin{tabular}{ c c c  c  c c  c  c c  c  c c  c  c c c}
\toprule[1.05pt]
Methods   & Pre   & Post   &\multicolumn{1}{c}{\begin{tabular}[c]{@{}l@{}}SE\end{tabular}} & \multicolumn{1}{c}{\begin{tabular}[c]{@{}l@{}}AS\end{tabular}} & \multicolumn{1}{l}{\begin{tabular}[c]{@{}l@{}}MA\end{tabular}} & \multicolumn{1}{c}{\begin{tabular}[c]{@{}l@{}}GSM\end{tabular}} & \multicolumn{1}{c}{Aqua}     & \multicolumn{1}{c}{Svamp}   & \multicolumn{1}{c}{\begin{tabular}[c]{@{}l@{}}DU\end{tabular}} & \multicolumn{1}{c}{\begin{tabular}[c]{@{}c@{}}TSO\end{tabular}} & \multicolumn{1}{c}{\begin{tabular}[c]{@{}l@{}}LLC\end{tabular}} & \multicolumn{1}{c}{\begin{tabular}[c]{@{}c@{}}CF\end{tabular}} & \multicolumn{1}{c}{\begin{tabular}[c]{@{}c@{}}CSQA\end{tabular}} & \multicolumn{1}{c}{\begin{tabular}[c]{@{}c@{}}SQA\end{tabular}} & \multicolumn{1}{c}{\begin{tabular}[c]{@{}l@{}}SQA-GLM4\end{tabular}}\\ 

\toprule[1.05pt]
Standard Finetuning&            &            &5.92&6.72&12.22&5.45&27.55&8.66&86.48&40.00&31.33&\textbf{100.00}&69.53&63.46 &63.46  \\
Pre-thinking       &$\checkmark$&            &3.29&2.52&6.10&6.07&15.74&4.33&82.88&\textbf{40.88}&21.33&\textbf{100.00}&49.63&58.22 &62.73 \\
Post-thinking      &            &$\checkmark$&6.58&\textbf{7.56}&12.22&5.91&28.74&8.00&83.78&38.66&28.00&\textbf{100.00}&70.59&64.91 &63.46  \\
ATM                &$\checkmark$&$\checkmark$&\textbf{7.24}&\textbf{7.56}&\textbf{16.11}&\textbf{6.26}&\textbf{29.92}&\textbf{10.00}&\textbf{87.38}&\textbf{40.88}&\textbf{33.33}&\textbf{100.00}&\textbf{70.67}&\textbf{65.35} &\textbf{64.77}                                              \\
\toprule[1.05pt]
\end{tabular}}
\vspace{-0.3cm}
\end{center}
\end{table*}

After \(c_i\) is obtained, the \(S\) can be split into pre-thinking dataset \(S_{pre}\) and post-thinking dataset \(S_{post}\). And then,  the following training objects can be utilized to train the perception module and generation module:
\begin{gather}
\mathbb{E}_{(x_i, r_i, y_i)\sim S_{pre}}[logP(r_i | x_i; \theta_s, \theta_p) \cdot logP(y_i | x_i, r_i,; \theta_s, \theta_p)]
\label{eq6}
\\
\mathbb{E}_{(x_i, r_i, y_i)\sim S_{post}}[logP(y_i | x_i; \theta_s, \theta_p) \cdot logP(r_i | x_i, y_i,; \theta_s, \theta_p)]
\label{eq7}
\end{gather}

With the above end-to-end training objects, the perception module will learn how to perceive the comlexity of questions so as to prompt the generation module to think ahead or later, and the generation module will learn how to answer the question with the soft prompt tokens.

The complete training process of the ATM is described in Algorithm \ref{algorithm}. Since most of the improvements made by other CoT-distillation methods on pre-thinking can be performed in step 2 and step 8, ATM can further improve the performance of these methods in a plug-and-play manner.

\section{Experiments}
\subsection{Experimental Setting}
\subsubsection{Datasets}
12 complex reasoning datasets divided into four classes are used to evaluate the proposed method: arithmetic (SingleEq \cite{singleeq}, AddSub \cite{addsub}, MultiArith \cite{mutilarith}, GSM8K \cite{gsm8k}, Aqua\cite{ling-etal-2017-program}, Svamp \cite{svamp}), symbolic \cite{NEURIPS2022_8bb0d291} (Last Letter Concatenation, Coin Flip), common sense (CommonSenseQA \cite{commonsenseqa}, StrategyQA \cite{strategyqa}) and other (Date Understanding, Track Shuffled Objects). The train-test splitting of these datasets follows ho et al.\cite{ho-etal-2023-large}. For the sake of simplicity, we use abbreviations (concatenated initials) to represent the name of each dataset in Table \ref{tab1}, Table \ref{tab2}, and Table \ref{tab: scal}.

\subsubsection{Models}
Unless otherwise stated, the teacher LLM refers to \textit{gpt-3.5-turbo} \cite{instructgpt} provided by OpenAI API. For student SLM, GPT2-Large \cite{gpt2} and T5-Large \cite{t5} are used to verify the effectiveness of ATM because not only are they representative models of decoder-only and encoder-decoder architectures respectively, but their parameter sizes also make them feasible in practice. Regardless of the teacher or student model, greedy decoding is applied when generating text.

\subsubsection{Training Details}
The student model is implemented by PyTorch \cite{pytorch} and trained on 2 NVIDIA 4090 GPUs. When training model, the batch size, initial learning rate and training epoch are set to 4, 1e-5 and 20 respectively. For varying learning rate, we apply cosine with restarts whose warm-up step is 1200. For optimizing model parameters, adam optimizer with \(\beta_1 = 0.9\), \(\beta_2 = 0.95\) and \(weight\_decay = 0.1\) is adopted. And the max rationale length is limited to 300. As for the \(K\), we set it to 50 across all tasks due to the better performance of this setting. 

\subsubsection{Evaluation Metric:} We apply accuracy
\begin{equation}
Acc= \frac {TP + TN} {FP + TP + FN + TN} \qquad
\end{equation}
to evaluate the performance of different methods, where $TP$, $FP$, $FN$ and $TN$ indicate true positive, false positive, false negative and true negative counts respectively.

\begin{table}
\caption{The performance of w./w.o. perception module and labeling assignment strategy. Accuracy (\%) is reported on StrategyQA and MultiArith using GPT2-Large.}
\begin{center}
\setlength\abovecaptionskip{-0.2cm}
\setlength\belowcaptionskip{-1cm} 
\label{tab3}
\renewcommand\arraystretch{1.5}
\begin{tabular}{ c c c}
\toprule[1.05pt]
Methods & StrategyQA &  Multi Arith\\
\toprule[1.05pt]
Base & 62.88 & 20.55\\
Base w. perception module & 63.31 &  21.11\\
Base w. labeling assignment strategy & 63.90 & 21.11 \\
ATM & 65.06 & 25.55\\
\toprule[1.05pt]
\end{tabular}
\vspace{-0.3cm}
\end{center}
\end{table}

\begin{figure}[tb]
    \centering
    \includegraphics[width=1\linewidth]{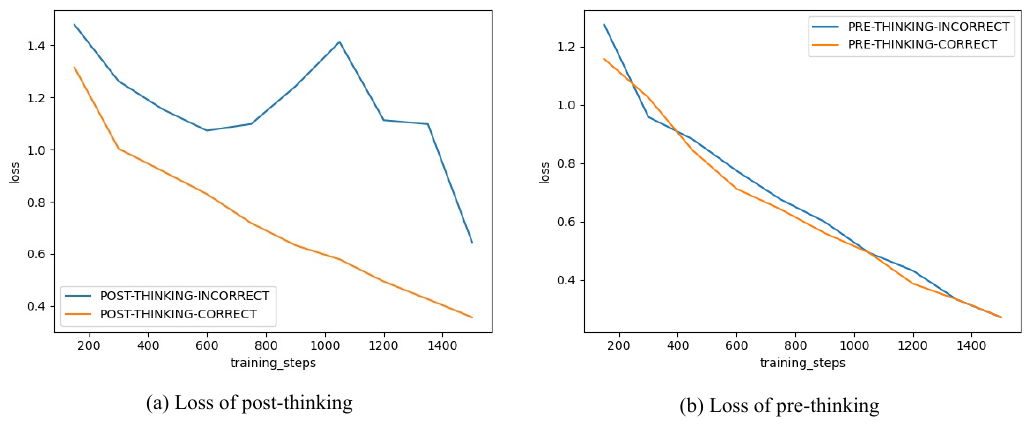}
    \caption{Loss variation against the training steps of the pre-thinking and post-thinking mechanism on different types of samples. The dataset and SLM are StrategyQA and GPT2-Large respectively.}
    \label{fig:loss}
\end{figure}

\begin{figure*}[tb]
    \centering
    \includegraphics[width=1\linewidth]{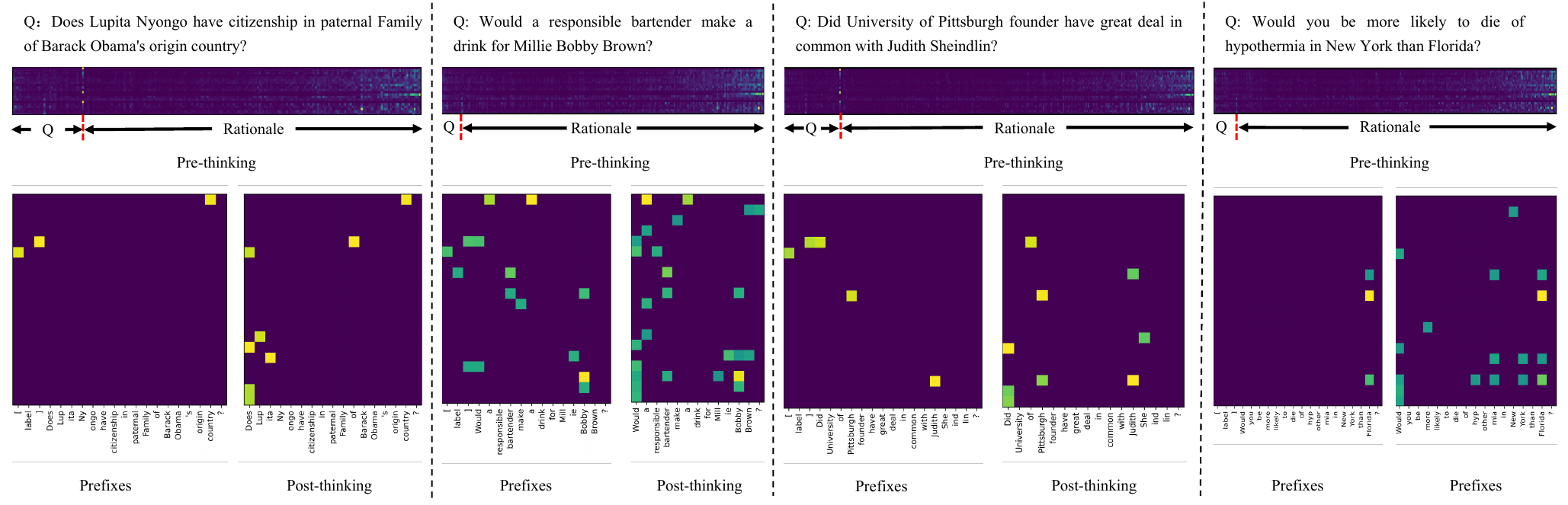}
    \caption{Attention visualization of GPT-large on StrategyQA with different CoT-finetuning strategies. The darker the color of a token is, the less attention will be paid to it in answer generation. Each row and column represents an attention head and a token, respectively. Since the rationale is too long, we omit the display of the specific question and rationale of pre-thinking. To compare the differences in attention among different methods more intuitively, all attention values that are less than 0.1 are set to zero in the attention map.}
    \label{figattn}
    \vspace{-0.3cm}
\end{figure*}

\subsection{Ablation experiment}
\subsubsection{Main result}
we evaluate the performance of the post-thinking and ATM across 12 reasoning tasks and two different SLMs (GPT2-Large and T5-Large) to verify the effectiveness of the two mechanisms. The results are reported in Table \ref{tab1} and Table \ref{tab2}. \textit{``Standard Finetuning''} in tables refers to the training object is:
\begin{gather}
\mathbb{E}_{(x_i, y_i)\sim S}[logP(y_i | x_i; \theta_s)]
\label{eq_s}
\end{gather}
Besides, it should be noted that the last column (StrategyQA-GLM4) in both tables indicates that the teacher LLM is GLM4 \cite{glm2024chatglm}. 

First, one can easily observe that post-thinking is superior to pre-thinking in most tasks, but performs slightly worse than pre-thinking on a few tasks (GSM8k in Table \ref{tab1} and Table \ref{tab2}, Svamp in Table \ref{tab1} and Shuffled Objects in Table \ref{tab2}), especially arithmetic tasks. This result shows that post-thinking is better than pre-thinking in tasks with low complexity while pre-thinking still has its irreplaceable advantages when faced with complex questions such as complex arithmetic questions that require thinking, reasoning, and breaking them down into sub-questions. Furthermore, ATM achieves better performance by further combining the reasoning capacity of pre-thinking and the ability of post-thinking to learn hard samples and avoid rationale-sensitive problem, which illustrates the quantitative effectiveness of ATM.

Secondly, by comparing the last two columns of Table \ref{tab1} and Table \ref{tab2}, it is evident that the proposed post-thinking and ATM mechanisms can enhance the performance of SLM under the guidance of various teacher LLM, further demonstrating the robustness of the proposed mechanism.

\subsubsection{Effectiveness of submodules in ATM}
in addition to the coarse-grained primary ablation experiments, fine-grained ablation studies are also conducted on two internal modules of the ATM, namely the perception module and the labeling assignment strategy. The results are presented in Table \ref{tab3}. \textit{``Base"} in Table \ref{tab3} refers to a naive implementation of the ATM mechanism, which means: 1) not utilizing the perception module, and 2) dividing the training dataset directly based on the correctness of the answers. In this context, when post-thinking model responses correctly, it indicates lower complexity of the input questions, and the samples are assigned to \(S_{post}\); otherwise, they are assigned to \(S_{pre}\).

It can be seen that the performance of \textit{``Base"} is not ideal, even performing worse than the results of standard finetuning in Table \ref{tab1}. We argue that this is mainly due to two reasons: 1) in the absence of other informational features, the same input format needs to correspond to both pre-thinking and post-thinking output forms, which increases the difficulty of model training; 2) in the naive labeling assignment strategy, due to common hallucination and other interfering factors in language models, directly using answer correctness as a hard criterion for dataset division introduces significant noise into the division results. Therefore, after adding the perception module and the labeling assignment strategy, which respectively introduce prompt information indicating problem complexity and employ a soft dataset division method by comparing the distribution differences between answer with rationale predicted by SLM and those generated by LLM, the model's performance improves. Finally, under the joint of the perception module and the labeling assignment module, the advantages of both post-thinking and pre-thinking are combined successfully, thus improving the performance significantly in the last row of Table \ref{tab3}.

\subsubsection{Effectiveness of ATM on more advanced SLMs} 
To explore whether the proposed method is also effective on more advanced SLMs, we additionally conduct experiments on two models: Qwen2.5-0.5B\cite{qwen2} and LLama3.2-1B\cite{grattafiori2024llama}. As shown in Table \ref{tab:betterSLM}, ATM improves performance on both models, demonstrating the generality and robustness of our approach.

\begin{table}[tb]
\begin{center}
\setlength\abovecaptionskip{-0.2cm}%用于调整标题前后的距离，可以自己设置来增加美观。  
\setlength\belowcaptionskip{-1cm} 
\caption{The performance of different strategies on Qwen2.5-0.5B and LLama3.2-1B. Accuracy (\%) is reported on StrategyQA.}
\renewcommand\arraystretch{1.5}
\resizebox{\linewidth}{15mm}{
\begin{tabular}{   c  c  c c  c }
\toprule[1.05pt]
Methods& Pre   & Post &     Qwen2.5-0.5B       &     LLama3.2-1B \\

\toprule[1.05pt]
Standard Finetuning  & &    & 64.77 & 64.91 \\
Pre-thinking        &$\checkmark $&     & 64.91 & 66.08 \\
Post-thinking      & & $\checkmark $    & 65.35 & 67.24 \\
ATM                &$\checkmark$ &  $\checkmark $   & \textbf{66.08} & \textbf{67.78}\\
\toprule[1.05pt]
\end{tabular}}
\vspace{-0.3cm}
\label{tab:betterSLM}
\end{center}
\end{table}

\begin{table*}[tb]
\vspace{-0.5cm} 
\begin{center}
\caption{The performance of different strategies on GPT2-Large and T5-Large. Accuracy (\%) is reported across 4 types of representative reasoning tasks.}
\renewcommand\arraystretch{1.5}
\renewcommand\arraystretch{1.5}
% \resizebox{\linewidth}{30mm}{
\begin{tabular}{ c  c  c  c c  c c c c c}
\toprule[1.05pt]
\multirow{2}{*}{Mechanism} & \multirow{2}{*}{Methods} & \multicolumn{2}{c}{MultiArith} & \multicolumn{2}{c}{Track Shuffled Objects} & \multicolumn{2}{c}{Last Letter Concatenation} & \multicolumn{2}{c}{StrategyQA} \\
                  &                   &    GPT2       &     T5     &    GPT2       &     T5      &     GPT2      &    T5      &      GPT2     & T5 \\

\toprule[1.05pt]
\multirow{8}{*}{Prompt Engineering}&CoT\cite{wei2022chain}       &0.00& 0.00 &0.00& 0.00 & 0.00 & 0.00  & 0.00 & 0.00 \\
&CoT w. \textit{gpt-3.5-turbo}       &\multicolumn{2}{c}{88.33} &\multicolumn{2}{c}{68.44}  & \multicolumn{2}{c}{64.66}   & \multicolumn{2}{c}{53.71} \\
&COD\cite{cod}       &0.00& 0.00 &0.00& 0.00 & 0.00 & 0.00  & 0.00 & 0.00 \\
&COD w. \textit{gpt-3.5-turbo}       &\multicolumn{2}{c}{82.22} &\multicolumn{2}{c}{28.88}  & \multicolumn{2}{c}{37.33}   & \multicolumn{2}{c}{54.14} \\
&SC\cite{wang2023selfconsistency}       &0.00& 0.00 &0.00& 0.00 & 0.00 & 0.00  & 0.00 & 0.00 \\
&SC w. \textit{gpt-3.5-turbo}       &\multicolumn{2}{c}{91.11} &\multicolumn{2}{c}{72.44}  & \multicolumn{2}{c}{68.00}   & \multicolumn{2}{c}{55.74} \\
&Self-Refine\cite{sr}       &0.00& 0.00 &0.00& 0.00 & 0.00 & 0.00  & 0.00 & 0.00 \\
&Self-Refine w. \textit{gpt-3.5-turbo}       &\multicolumn{2}{c}{81.66} &\multicolumn{2}{c}{-}  & \multicolumn{2}{c}{-}   & \multicolumn{2}{c}{-} \\
\hline
\multirow{4}{*}{Prefixes}&Distilling Step-by-Step\cite{hsieh-etal-2023-distilling}       &21.11&12.22 &41.33&40.44 & 5.33 & 26.00 & 63.61 & 64.33 \\
&Mint\cite{liang-etal-2024-mint-boosting}       & 23.88 & 12.22 & - &- & - & - & - &  - \\
&MMI\cite{MMI}       &-& 16.66 &-& 42.66 & - & 41.33  & - & 64.77\\
&Mind’s Mirror\cite{minds}  & 23.88  & 15.55 & 43.55 & 42.22 & 9.33 & 39.33 & 64.19 &   65.06  \\
\hline
Pre-thinking&Fine-tune-CoT \cite{ho-etal-2023-large}& 22.77 & 8.33 & 41.33  & 42.22 & 4.00 & 29.33 & 64.19 &  62.44  \\
Post-thinking&Fine-tune-CoT w. post-thinking & 23.33 & 13.88 & 44.44 &40.00 & 4.00 & 32.66 & 64.33 & 65.50                                             \\
ATM&Fine-tune-CoT w. ATM & 26.66 & 17.22 & 46.22 & 42.66 & 4.66 & 35.33 & 65.06 &  66.37                                            \\
\hline
Pre-thinking&SCOTT \cite{wang-etal-2023-scott} & 25.55 & 15.55 & 41.33 & 42.66 & 4.66 & 27.33 & 63.46 &  65.06 \\
Post-thinking&SCOTT w. post-thinking & 26.11 & 14.44 & 42.66 & 40.00 & 5.33 & 32.66 & 66.37 & 65.35\\
ATM&SCOTT w. ATM & 28.88 & 17.77 & 45.33 & 44.00 & 7.33 & 37.33 & 66.81 & 67.24\\
% \hline
% Pre-thinking&Turning Dust into Gold\cite{li2023turning} &  &  & 17.33 & &  &  &  & \\
% Post-thinking&Turning Dust into Gold w. post-thinking &  &  &  & &  &  &  & \\
% ATM&Turning Dust into Gold w. ATM &  &  &  & &  &  &  & \\
\hline
Pre-thinking&Specializing\cite{FuSpecializing} & 21.66 & 8.33 & 40.00 & 40.88 & 4.00 & 24.66 & 63.75 & 62.44\\
Post-thinking&Specializing w. post-thinking & 22.77 & 13.88 & 40.88 & 40.44 & 5.33 & 31.33 & 64.92 & 64.92\\
ATM&Specializing w. ATM & 25.55 & 16.66 & 42.66 & 41.77 & 5.33 & 34.66 & 65.50 & 65.79 \\
\hline
ATM & ATM &25.55 &16.11 &43.55 & 40.88 & 4.66 & 33.33 & 65.06 & 65.35 \\
\toprule[1.05pt]
\end{tabular}
\vspace{-0.3cm}
\label{tab:gpt2sota}
\end{center}
\end{table*}

% \usepackage{multirow}
\begin{table}[ht]
\caption{The efficiency and complexity of different strategies on GPT2-Large. Parameters (M), reasoning time (S/Sample), average accuracy (\%), training GFLOPs, and training time (Hours) are reported across 4 types of representative reasoning tasks (MA, TSO, LLC, and SQA). Vanilla post-thinking and vanilla ATM mean that no additional strategies are plugged in, while other post-thinking and ATM methods are combined with corresponding strategies.}
\renewcommand\arraystretch{1.5}
\resizebox{\linewidth}{24mm}{
\begin{tabular}{ccccccc}
\toprule[1.05pt]
Mechanism                      & Methods                 & Parameters & Reasoning Time & Accuracy & GFLOPs & Training Time \\ \toprule[1.05pt]
\multirow{3}{*}{Pre-thinking}  & Fine-tune-CoT \cite{ho-etal-2023-large}           & 774            & 9.8                      & 33.07   & 784.09                      & 7             \\
                               & SCOTT \cite{wang-etal-2023-scott}                & 774            & 8.9                      & 33.75    & 784.09                    & 4            \\
                               & Specializing  \cite{FuSpecializing}          & 774            & 8.8                      & 32.35  & 993.44                      & 7              \\ \hline
\multirow{2}{*}{Prefixes}      & Distilling Step-by-Step \cite{hsieh-etal-2023-distilling}  & 774            & 0.1                      & 32.84    & 1397.8                      & 3            \\
                               & Mind’s Mirror \cite{minds}          & 774            & 0.1                      & 35.23     & 1397.8                      & 3                 \\ \hline
\multirow{4}{*}{Post-thinking} & Vanilla                 & 774            & 0.1                      & 33.61    & 784.09                      & 2             \\
                               & Fine-tune-CoT \cite{ho-etal-2023-large}          & 774            & 0.1                      & 34.02     & 784.09                      & 7           \\
                               & SCOTT \cite{wang-etal-2023-scott}                   & 774            & 0.1                      & 35.11    & 784.09                     & 4             \\
                               & Specializing\cite{FuSpecializing}            & 774            & 0.1                      & 33.47     & 993.44                      & 7           \\ \hline
\multirow{4}{*}{ATM}           & Vanilla                 & 778            & 6.1                      & 34.70    & 784.09                      & 6              \\
                               & Fine-tune-CoT \cite{ho-etal-2023-large}           & 778            & 6.1                      & 35.65     & 784.09                      & 21                             \\
                               & SCOTT  \cite{wang-etal-2023-scott}                 & 778            & 6.0                      & 37.08   & 784.09                      & 12              \\
                               & Specializing  \cite{FuSpecializing}          & 778            & 6.2                      & 34.76    & 993.44                      & 22             \\ \toprule[1.05pt]
\end{tabular}}
\vspace{-0.3cm}
\label{tab: efficiency}
\end{table}

\begin{table}[]
\caption{The display about how the model (GPT-Large) trained with ATM to divide the thinking mode on the StrategyQA test set. 10 pre-thinking and 10 post-thinking questions were randomly selected, respectively.}
\begin{tabular}{p{0.95\linewidth}}
\toprule[1.05pt]

\multicolumn{1}{c}{Pre-thinking Questions} \\
\toprule[1.05pt]
1. In a hypothetical race between a Swallow and an American Woodcock, would the Swallow win?                      \\
                   2. Would a person with Anorexia nervosa be more likely to break a bone than a regular person?                      \\
                   3. Could a hundred thousand lolcats fit on a first generation iPhone?  \\
                   4. Would members of Blue Lives Matter support every element of Grand Theft Auto III?                      \\
                   5. Would a viewer of Monday Night Football be able to catch WWE Raw during commercial breaks?                      \\
                   6. Does ontology require a scalpel?                      \\
                   7. Could Eddie Murphy's children hypothetically fill a basketball court by themselves?                      \\
                   8. Can I find my home with latitude and longitude?                      \\
                   9. Can you substitute the pins in a bowling alley lane with Dustin Hoffman's Oscars?                      \\
                   10. Would you be likely to see storks at a baby shower?                      \\
\toprule[1.05pt]
\multicolumn{1}{c}{Post-thinking Questions} \\ \toprule[1.05pt]
1. Is Christmas celebrated during winter?                      \\
                   2. Is it safe to wear sandals in snow?                      \\
                   3. Do guitarists need both hands to play?                      \\
                   4. Are people banned from entering the Forbidden City?                      \\
                   5. Is Thanksgiving sometimes considered a day of mourning?                      \\
                   6. Would the United States Military Academy reject an applicant with multiple sclerosis?                      \\
                   7. Do you need to worry about Zika virus in Antarctica?                      \\
                   8. Did Dr. Seuss live a tragedy free life?                     \\
                   9. Can food be cooked in the cosmic microwave background?                      \\
                   10.Did Brazilian jiu-jitsu Gracie founders have at least a baker's dozen of kids between them?                      \\
\toprule[1.05pt]
\end{tabular}
\label{tab:q}
\end{table}

\subsection{Post-thinking vs. Pre-thinking}

\subsubsection{Loss analysis} First, in order to more intuitively analyze the effect of post-thinking's error amplifier function, we show the loss variation against the training steps of the pre-thinking and post-thinking mechanisms on different types of samples in Fig \ref{fig:loss}.  As shown in Fig \ref{fig:loss}, regardless of the training stage, in the pre-thinking mechanism, the loss gap between correctly and incorrectly predicted samples remains minimal. However, for post-thinking, as training progresses, the model gradually learns that the generated rationale needs to justify the predicted answer's validity. Consequently, when the answer prediction is incorrect, the rationale explains from the perspective of the wrong answer, causing the reasoning logic within the rationale to increasingly deviate from that needed to correctly answer the question. Therefore, the loss gap between correctly and incorrectly predicted samples begins to widen, which makes the model focus more on learning from incorrectly predicted samples, ultimately enhancing the model's performance. 
\subsubsection{Attention map}
Second, we visualize the attention map between the token that is used to generate answers and other tokens in Fig. \ref{figattn}. Through the attention map, one can first find that the answer generation overly relies on the rationale under the pre-thinking setting because almost all tokens with higher attention values are included in the rationale, which causes the correctness of the answer to be seriously affected even though just minor errors occur in rationale. In addition, the tokens with higher attention value in the question are tend to occur at the end of the question as shown in the position corresponding to the red vertical dotted line of the Fig. \ref{figattn}, which indicates that pre-thinking compresses the question information to the last token of the question so that the question information that the SLM can notice when generating the answer is lossy. Compared with pre-thinking, post-thinking can focus more on the complete information of the question itself when generating answers, and can escape from the direct influence of minor errors in rationale. Furthermore, it can be observed that the attention sparsity in the attention map from post-thinking to prefixes mechanism increases, indicating that post-thinking can notice more key points in the question when compared with prefixes mechanism.

\subsubsection{Efficiency}
Thirdly, we compare the inference speed between pre-thinking and post-thinking on three different datasets in Table \ref{tab: efficiency}. The inference speed of post-thinking is at least 50 times faster than that of pre-thinking, which is important for practical tasks that are sensitive to model inference time.

% \begin{figure}[tb]
%     % \setlength{\belowcaptionskip}{-0.5cm} %调整图片标题与下文距离
%     \centering
%     \includegraphics[width=1\linewidth]{imgs/fig_speed.pdf}
%     \caption{Inference speed of GPT2-Large on 3 datasets with different cot-finetuning strategies. As shown in the figure, since the post-thinking can stop inference after giving a precise answer, its speed is much faster than the compared pre-thinking method.}
%     \label{fig:speed}
% \end{figure}

\begin{figure*}[tb]
    \centering
    \includegraphics[width=0.9\linewidth]{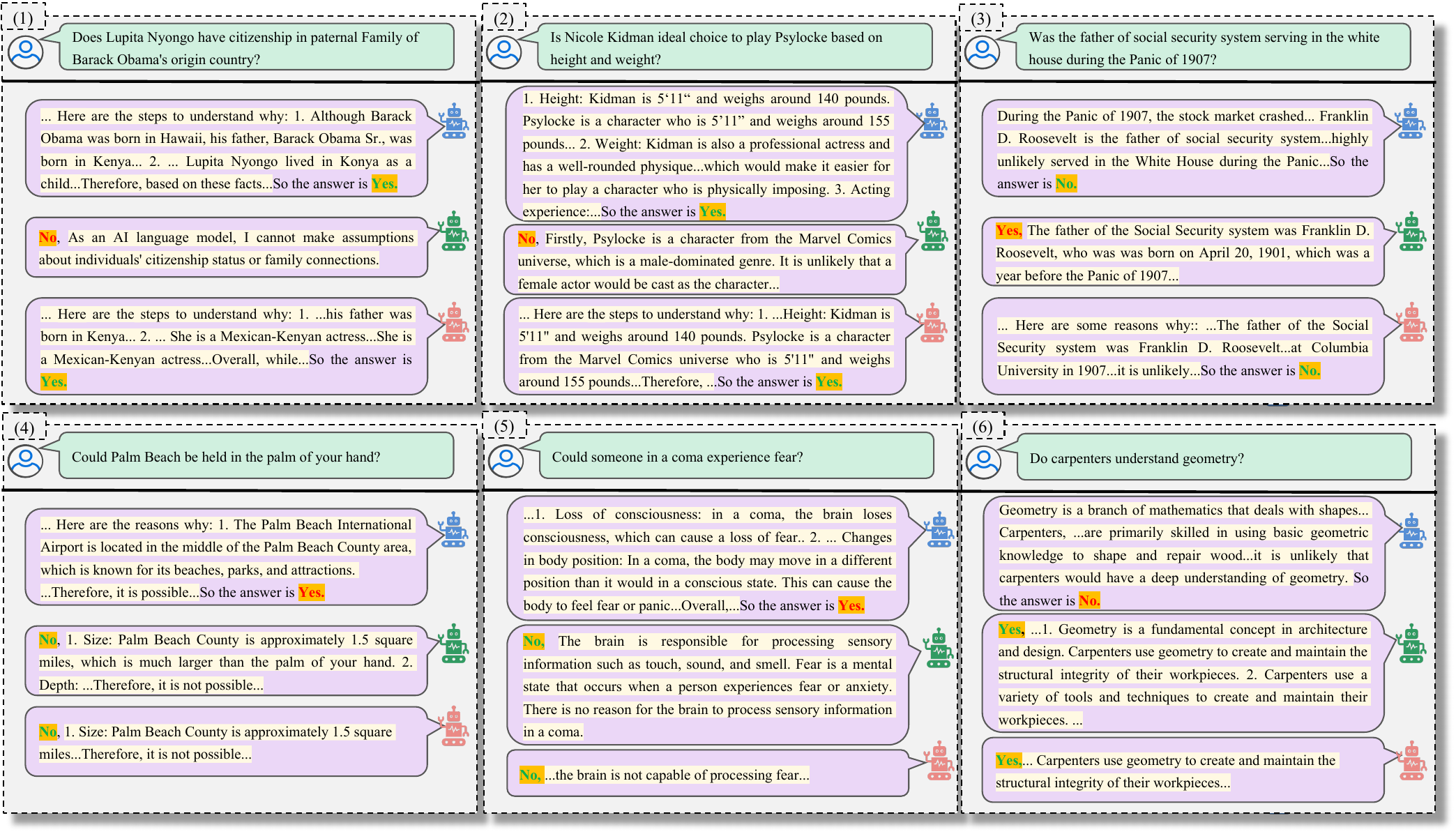}
    \caption{Case study. Six typical samples that can reflect the main characteristics of the pre-thinking, post-thinking and ATM are presented.}
    \label{fig:case}
    \vspace{-0.3cm}
\end{figure*}

\subsubsection{Case study}
finally, we analyze the samples where only one of the pre-thinking and post-thinking mechanisms predicts correctly to intuitively demonstrate their characteristics. Fig. \ref{fig:case} shows some representative samples.

In the samples (1, 2, and 3) correctly predicted by pre-thinking, one can observe that the types of questions are highly complex. Addressing them typically involves breaking down the original question into several sub-questions and then synthesizing the answers to each sub-question to provide a comprehensive response. For example, in sample 3, answering the question needs to infer who is the father of the social security system, and then deduce the institution he was serving in 1907. Such high-complexity questions are evidently not suitable for the post-thinking mechanism, causing post-thinking to output incorrect answers for these three questions. But on the other hand, we can intuitively grasp the error amplifier function of post-thinking from these three samples, i.e. when providing incorrect answers, post-thinking will explain from a peculiar perspective. For instance, in sample 2, to explain why Nicole Kidman is not an ideal choice to play Psylocke, post-thinking states that Psylocke is a male-dominated character (when in reality, Psylocke is a female character). Furthermore, in sample 1, post-thinking even chose not to explain directly. Although this error amplification effect has a limited impact on these highly complex questions, it can help the SLM focus on learning from hard samples that are low in complexity but challenging for the SLM to answer correctly, thereby improving the SLM's performance on these hard samples.

In the post-thinking correctly predicted samples (4, 5, 6), we can observe a relatively lower level of complexity in the questions. And for the pre-thinking mechanism, due to the absence of any constraints during the rationale generation process, the thought process can easily veer off in directions unrelated to answering the question, sometimes even leading in the wrong direction. For instance, the majority of the text consistently focuses on Palm Beach International Airport in the rationale of sample 4 and actions unrelated to answering the question (such as moving the body) are introduced in sample 5, resulting in erroneous thinking. This issue is evident even in pre-thinking correctly predicted samples. For example, in the rationale of sample 2, besides the perspective of height and weight, the pre-thinking self-generated an unrelated thinking direction (the acting experience). In contrast, for simpler samples, pre-thinking not only avoids the rationale-sensitive problem but also, guided by the correct answers, its rationale is more closely aligned with precise explanations. Based on the above qualitative analysis, we additionally calculated the BLEU scores for the rationales generated by post-thinking and pre-thinking compared to those generated by the LLM, resulting in scores of 33.56 and 32.12 respectively, which quantitatively further demonstrates the guiding and constraining effects of answer on rationale generation in post-thinking.

\subsection{ATM vs. (Post-thinking \& Pre-thinking)}
Fig. \ref{fig:case} also illustrates the ability of the ATM to integrate the strengths of both post-thinking and pre-thinking. Obviously, since ATM can perceive the complexity of the input question and form a soft prompt to indicate SLM whether to pre-think or post-think, it gets the correct answer in all six cases.  

In order to further evaluate the perception ability of ATM on the complexity of input questions, 10 questions are randomly selected from each of the two question sets used for pre-thinking and post-thinking respectively in ATM, and are shown in Table \ref{tab:q}. In the table, questions answered through pre-thinking are typically of higher complexity, often requiring SLM to decompose the questions, engage in reasoning, and perform comparative analyses to obtain answers. Conversely, questions answered using post-thinking in the table are of lower complexity and can often be directly resolved. Thus, this adaptive mechanism in ATM, which selects thinking approaches based on question complexity, is considered to be more flexible and closely aligned with human question-answering methods compared to pre-thinking and post-thinking. 

However, in Table \ref{tab:q}, from a human perspective, one can also observe that the set of questions using post-thinking contains several questions  (7, 9) that appear complex, while the other question set includes some questions (6, 8, 10) with lower complexity. This is related to the model's inherent question-solving ability. Influenced by factors such as training data and methods during the pre-training, different SLM possess varying foundational capabilities. Since ATM essentially categorize the question sets based on the SLM's own abilities, it leads to minor differences in question categorization between ATM and human perception of question complexity.

\subsection{Comparison with Other Methods}
We compare the performance and efficiency of our approach with 3 other types of state-of-the-art models related to ours on 4 datasets in Tables \ref{tab:gpt2sota} and \ref{tab: efficiency}. The first type of comparison method is the prefixes mechanism including distilling Step-by-Step \cite{hsieh-etal-2023-distilling}, Mint \cite{liang-etal-2024-mint-boosting}, MMI \cite{MMI}, and Mind’s Mirror \cite{minds}. \textit{`-'} of Mint and MMI in Table \ref{tab:gpt2sota} means that Mint cannot be applied to tasks other than arithmetic reasoning and MMI releases their implementation only based on the t5 framework, respectively. The second type of comparison method is the pre-thinking mechanism including Fine-tune-CoT \cite{ho-etal-2023-large}, SCOTT \cite{wang-etal-2023-scott} and Specializing \cite{FuSpecializing}. We integrate all the above methods into our unified training framework and data based on their open source implementations for fair comparison. Another type of comparison method is prompt engineering including CoT \cite{wei2022chain}, CoD \cite{cod}, SC \cite{wang2023selfconsistency}, and Self-Refine \cite{sr}. All methods of prompt engineering are tested on both LLM(w. \textit{gpt-3.5-turbo} in Table \ref{tab:gpt2sota}) and SLM. Furthermore, in the table, w. post-thinking refers to that transferring training object related to Equation \ref{eq1} in pre-thinking methods to \ref{eq2}, and w. ATM represents that the strategies proposed by the above pre-thinking methods are used in steps 2, 3, and 8 of Algorithms \ref{algorithm}, respectively.

\subsubsection{Accuracy} In Table \ref{tab:gpt2sota}, one can first observe that although prompt engineering enables LLM to achieve good performance on different reasoning tasks, they do not work on SLM, which proves the necessity of CoT-distillation. Second, ATM shows comparable performance to existing excellent distillation methods. Moreover, The plug-and-play improvement of post-thinking and ATM over pre-thinking methods is also demonstrated.
\subsubsection{Efficiency and Complexity} In Table \ref{tab: efficiency}, a comprehensive comparison of the efficiency and complexity of ATM and other methods are displayed. For the training phase, the perception module in ATM results in only a limited increase in overall space (parameters) and time (GFLOPs) complexity. And the overall training time increase as ATM requires training two additional models for the pre-thinking and post-thinking modes during data partitioning. In summary, although ATM incorporates additional components and strategies during training, the overall overhead remains within a reasonable range. As for the practical deployment of SLM, memory consumption and inference speed during testing are more critical. As shown in Table \ref{tab: efficiency}, ATM slightly increases memory usage due to the additional parameters introduced in the perception module. In terms of inference speed, ATM benefits from the post-thinking mechanism, making its overall inference speed significantly faster than pre-thinking-based methods. Among all methods, prefixes and post-thinking achieve the fastest inference speed. However, their overall performance is lower than ATM, and prefixes suffers from inconsistencies between the reasoning process and the final answer. In conclusion, ATM maintains the low-resource requirements of SLM while achieving the best trade-off between speed and accuracy.

Furthermore, we also compare ATM with other representative human-inspired dual-system thinking framework, comprising fast and slow thinking. We apply their official implementations to reproduce the result. As shown in Table \ref{tab: dual-system}, ATM outperforms these baselines, thereby further validating its effectiveness and advantages.

\begin{table}[]
\centering
\caption{Comparison of different human-inspired dual-system thinking frameworks. The student model is GPT2-Large and the accuracy (\%) of each dataset is reported. }
\renewcommand\arraystretch{1.25}
\begin{tabular}{ccccc}
\toprule[1pt]
Method        & MA & TSO & LLC & SQA  \\ \toprule[1pt]
Dualformer\cite{su2024dualformer}    & 23.88   & 43.11   & 4.00    &   64.19   \\
LS-Mixture\cite{yu2025long}    & 21.66   & 41.33   & 4.00    &   63.75   \\ \hline
ATM           & 25.55   & 43.55   & 4.66    &   65.06   \\ \toprule[1pt]
\end{tabular}
\vspace{-0.3cm}
\label{tab: dual-system}
\end{table}

\subsection{Discussion}

\begin{table}[]
\caption{A comparison of the quality of rationales generated by different teacher LLMs and their impact on the student SLM GPT2-Large.}
\renewcommand\arraystretch{1.5}
\resizebox{\linewidth}{8mm}{
\begin{tabular}{cccc}
\toprule[1.05pt]
Teacher        & Intermediate Errors(\%) & Other Errors (\%) & Accuracy of SLM (\%)\\ \toprule[1.05pt]
\textit{gpt-3.5-turbo}      & 40                  & 4            & 43.55           \\
Multi agent debate\cite{Multi-Agent-Debate} & 20                  & 0            & 49.78           \\ \toprule[1.05pt]
\end{tabular}}
\vspace{-0.3cm}
\label{tab: better teacher}
\end{table}

\begin{table}[]
\caption{Comparison of different distillation mechanisms on the unified cross-lingual and cross-task reasoning dataset. The overall accuracy (\%) and the accuracy (\%) of each subset of datasets are reported. }
\renewcommand\arraystretch{1.5}
\resizebox{\linewidth}{13.5mm}{
\begin{tabular}{ccccccc}
\toprule[1.05pt]
Method        & MA & TSO & LLC & SQA & CMATH & AVERAGE \\ \toprule[1.05pt]
Standard Finetuning            &  25.00  &  37.77   &  7.33   &  63.17   &   3.88    &   27.43      \\
Pre-thinking  &  33.33  &  46.22   &  21.33   &  60.26   &   3.33    &   32.89      \\
Post-thinking &  27.22  &  43.11   & 15.33    &  64.19   &   3.88    &   30.74      \\ \hline
ATM           &  35.55  &  49.77   &  22.66   &  65.21   &   4.44    &   35.52      \\ \toprule[1.05pt]
\end{tabular}}
\label{tab: scal}
\end{table}

\begin{table}[ht]
\centering
\caption{Accuracy (\%) and GFLOPs on MA with different features fed into perception module. The student model is GPT2-Large. $\checkmark$ represents that the corresponding feature is used.}
\renewcommand\arraystretch{1.5}
\resizebox{\linewidth}{12mm}{
\begin{tabular}{cccccccc}
\toprule[1.05pt]
Method & Query Tokens & Question Tokens & Word Count & Readability & MDD & Accuracy & GFLOPs\\ \toprule[1.05pt]
ATM    &      $\checkmark$        &                 &            &             &     & 23.88  &  790.65  \\
ATM    &      $\checkmark$        &        $\checkmark$         &            &             &     & 25.00 & 791.38   \\
ATM    &   $\checkmark$           &         $\checkmark$        &       $\checkmark$     &             &     & 25.55  & 791.43  \\
ATM    &    $\checkmark$          &       $\checkmark$          &    $\checkmark$        &     $\checkmark$        &     & 25.55   & 791.44\\
ATM    &    $\checkmark$          &   $\checkmark$              &   $\checkmark$         &  $\checkmark$           & $\checkmark$    & 26.11   & 791.48\\ \toprule[1.05pt]
\end{tabular}}
\vspace{-0.3cm}
\label{tab:additionalfea}
\end{table}

\begin{figure*}[tb]
    \centering
    \includegraphics[width=0.8\linewidth]{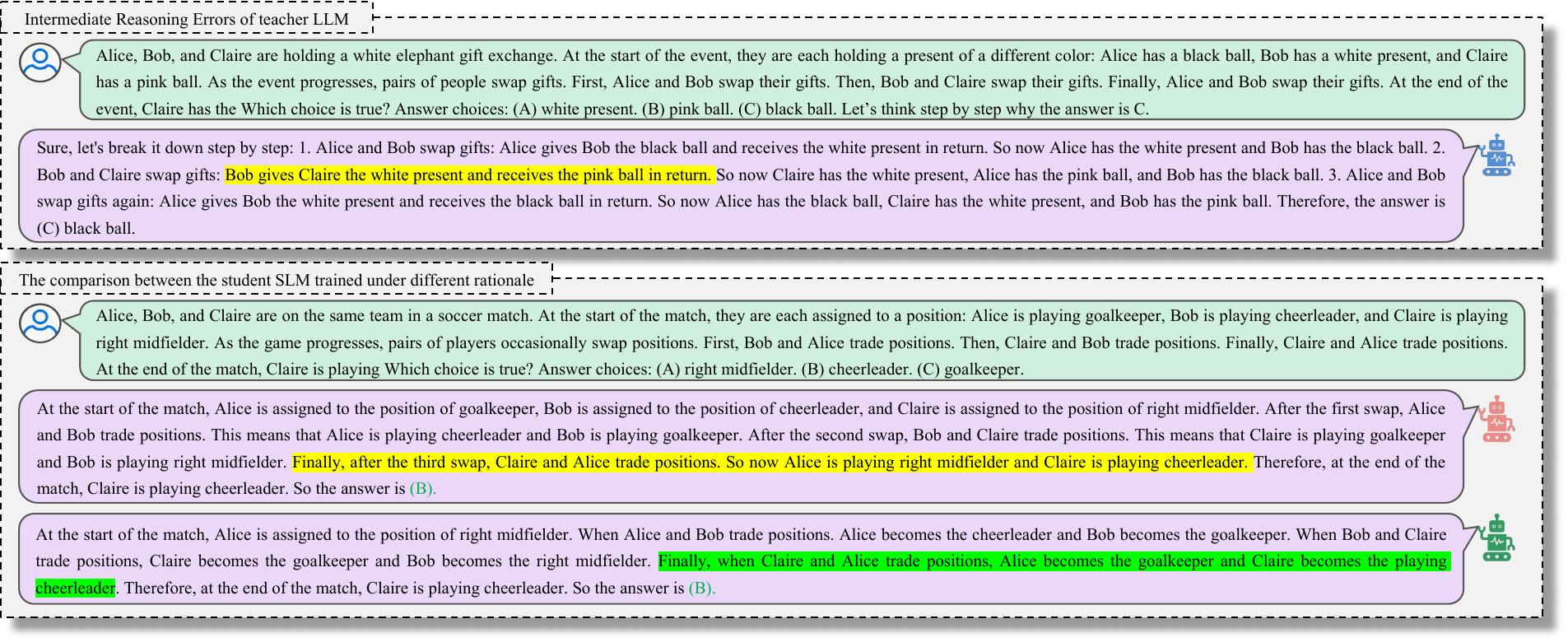}
    \caption{Case study on intermediate reasoning errors. Text highlighted in yellow indicates incorrect reasoning steps, while text highlighted in green represents correct intermediate steps. The green answer indicates that the answer is correct. The blue, red, and green robot icons correspond to the teacher LLM (\textit{gpt-3.5-turbo}), the SLM trained on rationales generated by \textit{gpt-3.5-turbo}, and the SLM trained on rationales generated through the multi-agent debate \cite{Multi-Agent-Debate}, respectively.}
    \label{fig:case_better_rationale}
\end{figure*}

\subsubsection{Better reasoners are better teachers} Although ATM incorporates answer information into a prompt to enhance the quality of the rationale generated by the teacher LLM, biased or suboptimal rationales are still inevitable. As shown in the first row of Table \ref{tab: better teacher}, we conduct a quantitative analysis on 50 randomly sampled cases from the TSO dataset. The results indicate that a primary issue in the LLM-generated rationale is minor errors in intermediate reasoning steps. Fig. \ref{fig:case_better_rationale} demonstrates this bias inherited by the student SLM from the teacher LLM. 

To mitigate this issues, integrating multiple teacher LLMs as a better reasoner is a feasible solution to reduce bias in the rationale. For an instance, we employ multi agent debate \cite{Multi-Agent-Debate} to aggregate the reasoning capabilities of multiple LLMs. In this setup, \textit{gpt-3.5-turbo} and GLM4 serve as the affirmative debater and negative debater, respectively, while \textit{gpt-4o-mini} acts as the judge. We then analyze the same 50 cases mentioned above, with the results presented in the second row of Table \ref{tab: better teacher}. It can be observed that errors in intermediate reasoning steps are significantly reduced under this multi-LLM framework. Subsequently, we use the improved rationale to train the SLM, and its performance is reported in Table \ref{tab: better teacher}. As we can see, the improvement in rationale quality leads to an increase in SLM performance. Moreover, a comparison of Fig. \ref{fig:case_better_rationale} illustrates a reduction in the influence of LLM bias on SLM.

Besides, a filtering strategy based on answer alignment is widely used to avoid the influence of low-quality teacher outputs. Specifically, one can employ regex matching for strings to compare the conclusion of rationale generated by the teacher LLM against the ground-truth answer, retaining only those samples where the answers align. Since correct conclusions correspond to the high-quality rationale, this filtering strategy significantly enhances the quality of the rationale used for distillation, thereby reducing the adverse effects of suboptimal teacher performance. Although this strategy may reduce the amount of available data, prior studies (e.g., Step-by-Step \cite{hsieh-etal-2023-distilling}) have demonstrated that CoT distillation remains effective even when only a subset of samples with high-quality rationale is preserved. This is because high-quality rationale enables the student model to get answers based on logical reasoning, rather than superficial pattern matching between questions and answers that rely on large datasets to learn. We validate that the proposed ATM exhibits this property, as shown in Table \ref{tab: filter}. Specifically, we simulated scenarios involving sub-performing teacher LLMs by varying the proportion of retained samples. The results indicate that even when only 50\% of the training data is retained, the student model still outperforms standard fine-tuning.

In summary, bias in the teacher LLM will affect the student SLM. Introducing a better reasoner has been proven to be an effective way to mitigate this issue, and the data filtering strategy play a significant role when the rationale of LLM is noisy.

\begin{table}[]
\centering
\caption{Effect of different training data rates on SQA. The rate refers to the proportion of data to be retained, and the student model is GPT2-Large. The accuracy (\%) is reported. }
\renewcommand\arraystretch{1.25}
\resizebox{\linewidth}{4mm}{
\begin{tabular}{cccccc}
\toprule[1pt]
Standard Finetuning    & ATM (rate=100\%) & ATM (rate=75\%) & ATM (rate=50\%) & ATM (rate=25\%) & ATM (rate=12.5\%)  \\ \toprule[1pt]
63.31                  &  65.06      &     64.33       &     64.63       &  62.73 &  60.26 \\ \toprule[1pt]
\end{tabular}}
\vspace{-0.5cm}
\label{tab: filter}
\end{table}

\subsubsection{Scalability of ATM} The scalability of ATM depends on whether the perception module can be effective across different applications and languages. In the perception module in ATM, query tokens pass through a cross-attention block, which makes query tokens interact with both the question and additional extracted features $a_i$ before being output. As a result, ATM can dynamically adjust the generation of soft prompt tokens based on the characteristics of the input question. This means that as long as the training data includes samples from different applications or languages, the perception module can learn to adjust soft prompt tokens accordingly. Consequently, the ATM can be extended to various applications and languages theoretically.
To further validate the scalability of soft prompt tuning in ATM, we construct a comprehensive cross-lingual and cross-task reasoning dataset, consisting of CMATH \cite{wei2023cmath} (Chinese mathematical reasoning), MA (English mathematical reasoning), SQA (English commonsense reasoning), LLC (English symbolic reasoning), and TSO (English logical reasoning). The dataset is split into training and validation sets with a 7:3 ratio. The results in Table \ref{tab: scal} demonstrate that ATM consistently outperforms other approaches, proving its scalability.
\subsubsection{Impact of features fed into the perception module} Whether incorporating a diverse set of explicit features in $a_i$ in addition to word count can enhance the learning capability of the perception module is displayed in Table \ref{tab:additionalfea}. In the table, readability is the perplexity of the input question calculated by GPT2-Large, while MDD is the Mean Dependency Distance of the input question. Readability and MDD are input into the model in the form of “\textit{Perplexity: num}” and “\textit{Mean Dependency Distance: num}”, respectively.

The results in Table \ref{tab:additionalfea} indicate that: a) Feeding question tokens into the perception module provides the most significant performance gain for SLM; b) As more explicit features are incorporated into $a_i$, the model’s performance gradually improves, but the observed improvements are relatively minor; c) The increase in model complexity resulting from these features is also minimal. These findings suggest that the perception module primarily relies on the implicit, comprehensive features of the question itself rather than $a_i$ for complexity estimation. If computational resources allow, the inclusion of carefully designed explicit features can offer measurable benefits. Conversely, in scenarios that prioritize model simplicity, this component can be entirely omitted without causing significant performance degradation.

\section{Conclusion}
In this work, we propose a plug-and-play adaptive thinking mechanism to distill the reasoning capacity from LLM to SLM for reasoning tasks, thereby improving the performance of SLM towards a specific task. By utilizing the perception module based on soft prompt tuning and the label assignment mechanism, ATM not only inherits the reasoning advantage of pre-thinking for complex questions but also introduces a post-thinking mechanism to alleviate the rationale-sensitive problem and strengthen the learning of hard samples. Extensive experiments on 12 reasoning datasets and two different language models verify the superiority of ATM.

\section{ACKNOWLEDGMENT}
This work was supported by the National Science Fund for Distinguished Young Scholars of China (No. 62325604), and the National Natural Science Foundation of China (No. 62276271 and No. 62421002).

\bibliographystyle{IEEEtran} %
\bibliography{custom} %

\vfill

\end{document}